\documentclass[10pt,twocolumn,letterpaper]{article}
\usepackage{authblk}
\usepackage{cvpr}              

\usepackage{comment}

\definecolor{cvprblue}{rgb}{0.21,0.49,0.74}
\usepackage[pagebackref,breaklinks,colorlinks,allcolors=cvprblue]{hyperref}
\usepackage{multirow}
\usepackage[table]{xcolor}
\def\paperID{10266} 
\def\confName{CVPR}
\def\confYear{2026}

\title{VGGT-360: Geometry-Consistent Zero-Shot Panoramic Depth Estimation}

\author{Jiayi Yuan$^{1}$, Haobo Jiang$^{2}$, De Wen Soh$^{1}$, Na Zhao$^{*1}$}
\affil{{$^1$ Singapore University of Technology and Design, $^2$ Nanyang Technological University}}
\affil{\tt\small jiayi\_yuan@mymail.sutd.edu.sg, \tt\small haobo.jiang@ntu.edu.sg, \tt\small dewen\_soh@sutd.edu.sg, \tt\small na\_zhao@sutd.edu.sg 
  }

\begin{document}
\maketitle
{
\renewcommand\thefootnote{}
\footnotetext{* Corresponding author.}
}
\begin{abstract}
This paper presents VGGT-360, a novel training-free framework for zero-shot, geometry-consistent panoramic depth estimation. Unlike prior view-independent training-free approaches, VGGT-360 reformulates the task as panoramic reprojection over multi-view reconstructed 3D models by leveraging the intrinsic 3D consistency of VGGT-like foundation models, thereby unifying fragmented per-view reasoning into a coherent panoramic understanding. To achieve robust and accurate estimation, VGGT-360 integrates three plug-and-play modules that form a unified panorama-to-3D-to-depth framework: 
(i) \textit{Uncertainty-guided adaptive projection} slices panoramas into perspective views to bridge the domain gap between panoramic inputs and VGGT’s perspective prior. It estimates gradient-based uncertainty to allocate denser views to geometry-poor regions, yielding geometry-informative inputs for VGGT. 
(ii) \textit{Structure-saliency enhanced attention} strengthens VGGT’s robustness during 3D reconstruction by injecting structure-aware confidence into its attention layers, guiding focus toward geometrically reliable regions and enhancing cross-view coherence. (iii) \textit{Correlation-weighted 3D model correction} refines the reconstructed 3D model by reweighting overlapping points using attention-inferred correlation scores, providing a consistent geometric basis for accurate panoramic reprojection.
Extensive experiments show that VGGT-360 outperforms both trained and training-free state-of-the-art methods across multiple resolutions and diverse indoor and outdoor datasets. The code is available at https://github.com/Yuanjiayii/VGGT-360.

\end{abstract}    
\section{Introduction}
\label{sec:intro}
Monocular depth estimation (MDE) from $360^\circ$ panoramic images is an emerging research problem that plays a critical role in omnidirectional perception.
Unlike conventional perspective depth, panoramic depth provides complete geometric context, which is essential for applications such as SLAM~\cite{zhang2021panoramic, jiang2023robust, jiang2021sampling}, virtual reality~\cite{huang20176, attal2020matryodshka, jiang2023se}, and autonomous navigation systems~\cite{deng2025omnistereo}.
However, panoramic MDE faces two major challenges:
\textbf{i)} Panoramas are typically represented by the equirectangular projection (ERP), which flattens a sphere into a 2D image and inevitably introduces severe geometric distortions, thereby making conventional perspective MDE models unsuitable for direct application~\cite{eder2019convolutions}; and
\textbf{ii)} The acquisition of large-scale annotated panoramic datasets is extremely difficult and costly, which severely limits the performance and generalization capability of existing supervised approaches~\cite{rey2022360monodepth, wang2024depth}.

\begin{figure}[t]
  \centering
  \includegraphics[width=1\linewidth]{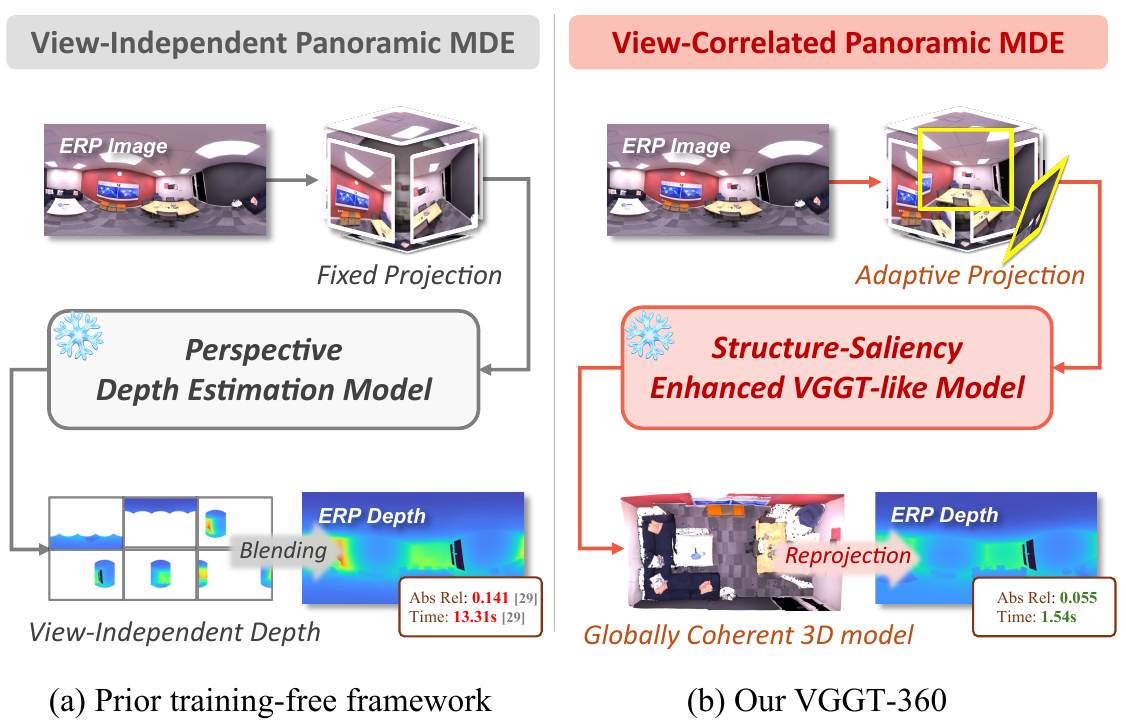}
  \vspace{-7mm}
  \caption{\textbf{Comparison between the conventional training-free panoramic depth estimation framework and our VGGT-360}. Unlike view-independent inference methods (\eg, 360MD~\cite{rey2022360monodepth}), VGGT-360 reconstructs a globally coherent 3D representation via VGGT-like 3D foundation models and reprojects it to the panorama, unifying fragmented per-view predictions into consistent, cross-view correlated depth with superior performance. 
}
\vspace{-4mm}
\label{Framework0}
\end{figure}

Existing panoramic MDE methods can be broadly categorized into \textit{training-based} methods and \textit{training-free} methods. 
Training-based approaches~\cite{yan2023distortion,yun2023egformer,cao2024crf360d,zhang2025sgformer} predict depth directly from ERP panoramas using sphere- or distortion-aware MDE models. However, their performance is limited by the scarcity of labeled panoramic data, resulting in lower accuracy and poor generalization. 
To address this, training-free methods~\cite{rey2022360monodepth, peng2023high, jung2025rpg360} decompose the panorama into perspective views, infer depth independently via pre-trained perspective MDE models, and fuse them into the full ERP map (Fig.~$\ref{Framework0} (a)$). 
Nonetheless, this strategy suffers from geometric inconsistency, as view-independent inference lacks cross-view interaction, leading to scale ambiguity and depth discontinuities across views, which in turn degrade geometric fidelity and structural detail.

In this paper, beyond the conventional view-independent, training-free paradigm for panoramic MDE, we reformulate the task as panoramic reprojection over multi-view reconstructed, globally consistent 3D models, as illustrated in Fig.~$\ref{Framework0}(b)$. 
Inspired by the success of VGGT~\cite{wang2025vggt} in capturing intrinsic global geometry across multiple views, we introduce \textbf{\textit{VGGT-360}}, a training-free, geometry-consistent panoramic MDE framework that leverages the holistic 3D reasoning capabilities of VGGT-like 3D foundation models to unify fragmented per-view reasoning into a coherent panoramic understanding.
However, since VGGT-like models (\eg, VGGT~\cite{wang2025vggt}, $\pi^3$~\cite{wang2025pi}) are originally trained on perspective images, directly extending them to panoramic scenes introduces a severe domain gap, making effective adaptation non-trivial. The network must simultaneously handle spherical distortion, non-uniform resolution, and $360^\circ$ wrap-around continuity, while producing geometrically consistent depth under unseen panorama scenarios.

To address these limitations, we innovatively design VGGT-360 as a three-stage, training-free framework comprising \textbf{i)}  uncertainty-guided adaptive projection, \textbf{ii)} structure-saliency enhanced attention, and \textbf{iii)} correlation-weighted 3D correction, jointly improving geometric fidelity and reconstruction robustness.
Specifically, \textit{uncertainty-guided adaptive projection} introduces gradient-derived geometric uncertainty to adaptively slice the panorama into perspective views. 
Unlike fixed projection schemes \cite{rey2022360monodepth, jung2025rpg360}, our adaptive method allocates denser sampling to geometry-deficient regions under coverage-and-overlap constraints, yielding more reliable, geometry-informative multi-view inputs for VGGT.
Furthermore, despite the remarkable zero-shot generalization of VGGT-like foundation models, their performance often degrades in weakly structured areas, compromising reconstruction quality.
To mitigate this, \textit{structure-saliency enhanced attention} injects a structure-saliency confidence map into VGGT’s attention layers, 
guiding the model toward reliable geometric and boundary cues for stable 3D reasoning.
Finally, \textit{correlation-weighted 3D model correction} refines the reconstructed model by assigning correlation-based reliability scores to overlapping 3D points, reinforcing consistent structures while suppressing ambiguous ones.

Notably, our VGGT-360 is highly general and plug-and-play, enabling seamless integration with diverse VGGT variants such as $\pi^3$~\cite{wang2025pi} and Fastvggt~\cite{shen2025fastvggt} for enhanced performance. Benefiting from the strong generalization of VGGT-like foundation models, VGGT-360 also exhibits remarkable zero-shot capability, delivering reliable depth estimation across varying resolutions and both indoor and outdoor scenes without any fine-tuning.

\noindent Our main contributions are summarized as follows: 

\begin{itemize}
    \item We propose \textbf{\textit{VGGT-360}}, a novel training-free, 3D model-aware framework that exploits the global consistency of VGGT-like 3D foundation models for coherent panoramic depth estimation, significantly outperforming prior view-independent panoramic MDE approaches. 
    \item We effectively adapt VGGT-like models for panorama-driven 3D reconstruction via three novel plug-and-play modules, \ie, uncertainty-guided adaptive projection, structure-saliency enhanced attention, and correlation-weighted 3D model correction, together enabling robust and structurally consistent panoramic depth prediction.
    \item Extensive experiments across multiple resolutions and diverse indoor/outdoor datasets validate the effectiveness and zero-shot generalization ability of VGGT-360. Notably, it surpasses previous SOTA methods by 27–36\% in Abs Rel on Stanford2D3D~\cite{armeni2017joint} and Replica360-2K~\cite{straub2019replica}.
\end{itemize}

\section{Related Work}

\noindent\textbf{Panoramic Depth Estimation.} 
Monocular depth estimation (MDE) for $360^\circ$ panoramic images aims to recover global scene depth from equirectangular projection (ERP) images. Existing learning-based methods mainly follow two directions. One designs sphere-aware networks that operate directly in the ERP domain to handle latitude-dependent distortions via distortion-aware convolutions or spherical positional encodings~\cite{tateno2018distortion, eder2019convolutions, sun2021hohonet, zhuang2022acdnet, yun2023egformer, cao2024crf360d, zhang2025sgformer}. The other adopts multi-projection fusion frameworks, which decompose panoramas into multiple perspective views to complement ERP processing and alleviate projection distortions~\cite{wang2018self, wang2020bifuse, jiang2021unifuse, li2022omnifusion, wang2022bifuse++, ai2023hrdfuse, mohadikar2024ms360}. However, both categories rely on supervised learning and are fundamentally limited by the scarcity of annotated panoramic datasets, leading to weak generalization in diverse scenes~\cite{wang2024depth, jung2025rpg360}.

\begin{figure*}[t]
 
  \centering
  \includegraphics[width=1\linewidth]{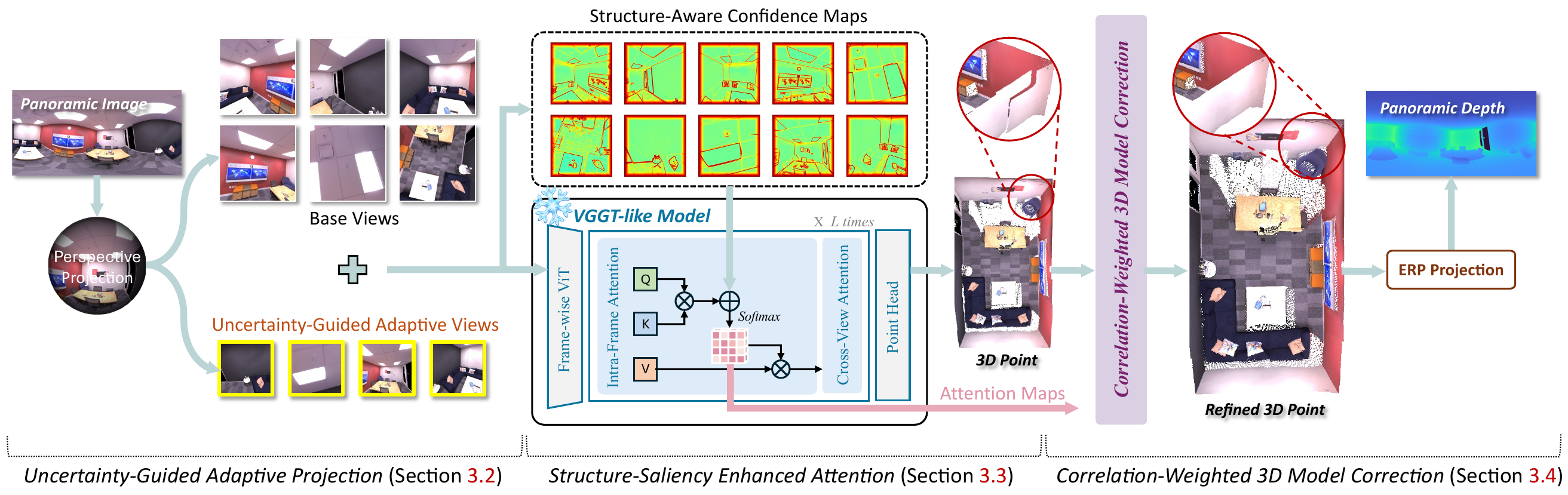}
  \vspace{-6mm}
  \caption{\textbf{Framework Overview of VGGT-360}. Given a panoramic image, we first perform \textit{uncertainty-guided adaptive projection} to produce geometry-informative views for VGGT. With \textit{structure-saliency enhanced attention}, VGGT reconstructs a structure-faithful 3D model, which is then refined by \textit{correlation-weighted 3D model correction} and reprojected into a globally consistent panoramic depth map.
}
  \vspace{-4mm}
\label{Framework}
\end{figure*}

\noindent\textbf{Zero-Shot Monocular Depth Estimation.} 
Recent monocular depth foundation models, such as MiDaS~\cite{ranftl2020towards}, Depth Anything~\cite{yang2024depth}, Metric3D~\cite{yin2023metric3d}, and Omnidata~\cite{eftekhar2021omnidata}, have demonstrated strong zero-shot generalization to unseen domains and scenes without fine-tuning.
Inspired by this, recent panoramic MDE approaches~\cite{rey2022360monodepth, peng2023high, wang2024depth, jung2025rpg360} have adopted these models as powerful pre-trained depth priors.
For instance, 360MD~\cite{rey2022360monodepth} first applied pre-trained perspective models to panoramic inputs by slicing the panorama into multiple views and blending their predictions through optimization-based fusion. Depth Anywhere~\cite{wang2024depth} and RPG360~\cite{jung2025rpg360} further introduced distillation and geometric optimization to enhance panoramic integration.
However, these zero-shot frameworks typically infer each view independently without enforcing cross-view geometric consistency, resulting in scale discrepancies and fragmented 3D structures.
This limitation motivates a training-free yet geometry-consistent framework that transforms fragmented per-view predictions into a unified panoramic depth map.

\noindent\textbf{Visual Geometry Grounded Transformer.} 
VGGT~\cite{wang2025vggt} is a unified architecture for reconstructing 3D models from multi-view images.
Unlike conventional depth foundation models~\cite{eftekhar2021omnidata, bhat2023zoedepth, yang2024depth, cao2025panda}, which learn implicit geometric priors from large-scale 2D datasets, VGGT explicitly reconstructs consistent 3D representations by leveraging cross-view geometric cues.
Subsequent VGGT variants~\cite{shen2025fastvggt, vuong2025improving, wang2025pi} have further enhanced its efficiency and robustness. For example, Fastvggt~\cite{shen2025fastvggt} accelerates inference with lightweight attention, while $\pi^{3}$~\cite{wang2025pi} eliminates reference-view dependency to enable permutation-equivariant reconstruction. Other extensions adopt VGGT as a geometry-aware prior for dense novel view synthesis~\cite{liu2025vggt} and robotic perception~\cite{ge2025vggt}.
In this work, we are the first to extend VGGT-like foundation models to panoramic depth, marking a significant step toward training-free omnidirectional 3D reasoning.

\section{Method}
In this paper, we present \textbf{\textit{VGGT-360}}, a novel training-free framework for geometry-consistent panoramic depth estimation.
As illustrated in Fig.~\ref{Framework}, given a single equirectangular panorama $\mathcal{I}_{\mathrm{erp}}{\in}\mathbb{R}^{H\times W\times3}$, VGGT-360 adaptively projects it into multiple perspective views, performs multi-view 3D reasoning through VGGT-like 3D foundation models (\eg, VGGT~\cite{wang2025vggt}, $\pi^3$~\cite{wang2025pi}, and Fastvggt~\cite{shen2025fastvggt}) to reconstruct a globally coherent 3D representation, and finally reprojects it back to the panorama to produce the depth map $\mathcal{D}_{\mathrm{erp}}{\in}\mathbb{R}^{H\times W}$ via ERP projection.

To realize this process effectively, VGGT-360 comprises three key training-free modules that enhance the performance of the VGGT-like model for panoramic input:
\textbf{i)} \textbf{Uncertainty-guided adaptive projection}, which generates perspective images for VGGT by allocating denser sampling to geometrically uncertain regions while preserving global spherical continuity (Section~\ref{UAP});
\textbf{ii) Structure-saliency enhanced attention}, which enhances VGGT’s attention by emphasizing structurally consistent geometric features, thereby improving 3D reasoning and suppressing artifacts in geometrically ambiguous regions (Section~\ref{SEA});
and \textbf{iii)} \textbf{Correlation-weighted 3D model correction}, which refines the 3D model by reweighting points in overlapping regions based on correlation cues derived from VGGT’s inter-frame attention, ensuring high-quality panoramic depth estimation (Section~\ref{CMC}).

\subsection{\textbf{VGGT-like 3D Foundation Models}} 
To enable multi-view reasoning, our framework leverages VGGT-like 3D Foundation Models as a core component. Given multi-view perspective inputs $\mathcal{V}_{\mathrm{per}}$, VGGT-like models tokenize images into patch embeddings and apply alternating intra-frame and cross-view attention for 3D reasoning, directly predicting camera parameters, point maps, depth, and point tracks. We adopt the point head output and reproject it into panoramic depth via ERP projection.

\subsection{Uncertainty-Guided Adaptive Projection}
\label{UAP}
To provide the perspective multi-view inputs required by VGGT, we first slice the equirectangular panoramic image $\mathcal{I}_{\mathrm{erp}}$ into a set of perspective views $v\in \mathcal{V}_{\mathrm{per}}$. Prior methods~\cite{rey2022360monodepth, wang2020bifuse, jung2025rpg360} typically adopt uniform, pre-defined projection schemes (\eg, cubemap), under the assumption that all viewing directions are treated equally, regardless of their geometric informativeness. In practice, views dominated by weakly structured regions (\eg, planar walls or ceilings) offer limited geometric cues, making depth reasoning more challenging. 
To address this, we propose \textit{uncertainty-guided adaptive projection}, which dynamically allocates denser views to regions with high geometric ambiguity, enabling VGGT to enrich its geometric understanding via neighboring views.
Our projection strategy consists of two stages: \textit{uncertainty-guided scoring}, which quantifies geometric ambiguity across base views, and \textit{adaptive neighbor augmentation}, which allocates additional view samples based on these uncertainties.

\begin{figure}[t]
  \centering
  \includegraphics[width=1\linewidth]{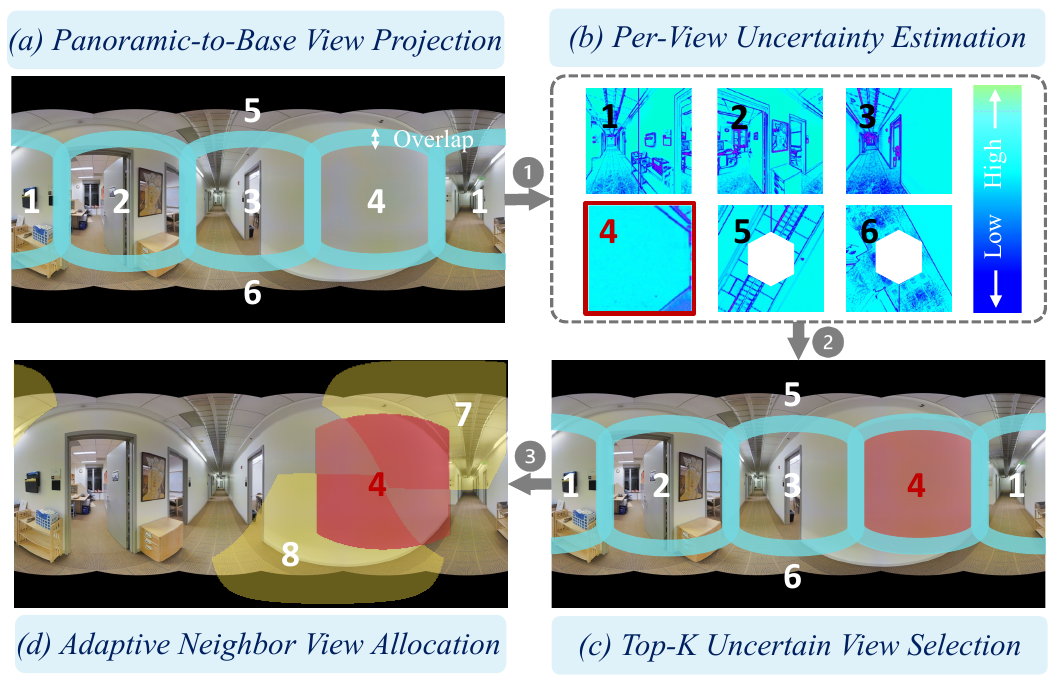}
  \vspace{-6mm}
\caption{
\textbf{Pipeline of our \textit{uncertainty-guided adaptive projection}}. 
We first generate $N_\mathcal{B}$ base views from the panorama, compute per-view uncertainty maps via edge-based scoring, and select the top-$K$ most uncertain views (with $N_\mathcal{B}{=}6$, $K{=}1$ in this example). These views are then augmented with neighboring projections to form a geometry-aware multi-view set as input for VGGT.
}
\vspace{-4mm}
\label{Fig3}
\end{figure}

\noindent{\textbf{Uncertainty-Guided Scoring.}} As shown in Fig.~$\ref{Fig3}(a$-$b)$, we begin with a pre-defined perspective projection using $N_\mathcal{B}$ ($N_\mathcal{B}\ge 6$) base views $\mathcal{B}{=}\left \{v_b^n\right \}_{n=1}^{N_\mathcal{B}}$, ensuring full coverage with controlled overlap. 
To identify geometry-sparse views for subsequent denser allocation, we assign each view an uncertainty score reflecting its geometric informativeness.
Given that gradient magnitude effectively captures edge richness and geometric saliency, we compute a per-pixel \textit{uncertainty map} $\mathbf{U}(p)=\sigma\left(-\mathbf{Z}(p)\right)$ from the grayscale image of each base view using the Sobel operator:
\begin{equation}
\mathbf{Z}(p){=}\left ({\mathbf{G}(p)-\operatorname{median}_{p'\in\Omega(v_b)}(\mathbf{G}(p'))}\right ) /{\tau},
\label{eq1}
\end{equation}
where $\mathbf{G}(p)$ denotes the Sobel gradient magnitude at pixel $p$, $\tau$ controls normalization sensitivity, and $\Omega(v_b)$ represents the valid region of base view $v_b$.
The area-weighted uncertainty score of view $v_b$ is computed as:
\begin{equation}
\small
\mathbf{S}(v_b)=
\frac{\displaystyle \sum_{p\in\Omega(v_b)} \mathbf{1}_{\mathrm{valid}}(p)\, \mathbf{U}(p)}
{\displaystyle \sum_{p\in\Omega(v_b)} \mathbf{1}_{\mathrm{valid}}(p)},
\end{equation}
where $\mathbf{1}_{\mathrm{valid}}(p)$ is an indicator function that masks out invalid pixels in perspective views.

\noindent\textbf{Adaptive Neighbor Augmentation.}
As shown in Fig.~$\ref{Fig3} (c$-$d)$, given the set of uncertainty scores $\{\mathbf{S}(v_b)\}_{v_b\in\mathcal{B}}$, we perform adaptive neighbor-augmented sampling by selecting 
the top-$K$ base views with the highest uncertainty:
$\mathcal{B}^*=\operatorname{Top-\textit{K}}\!\left(\{\mathbf{S}(v_b)\}_{v_b\in\mathcal{B}}\right)$.
For each selected view $v_b^*\in\mathcal{B}^*$, we generate two neighboring views $\mathcal{N}(v_b^*)$ with predefined yaw and pitch offsets (\ie, upper-right and lower-left of the view center).
This adaptive strategy yields an efficient perspective view set: $\mathcal{V}_{\mathrm{per}}{=}\mathcal{B}\cup\mathcal{N}(\mathcal{B}^*)$,
which maintains global coverage while adaptively increasing sampling density in geometrically ambiguous regions under a limited budget.

\begin{figure}[t]
  \centering
  \includegraphics[width=1\linewidth]{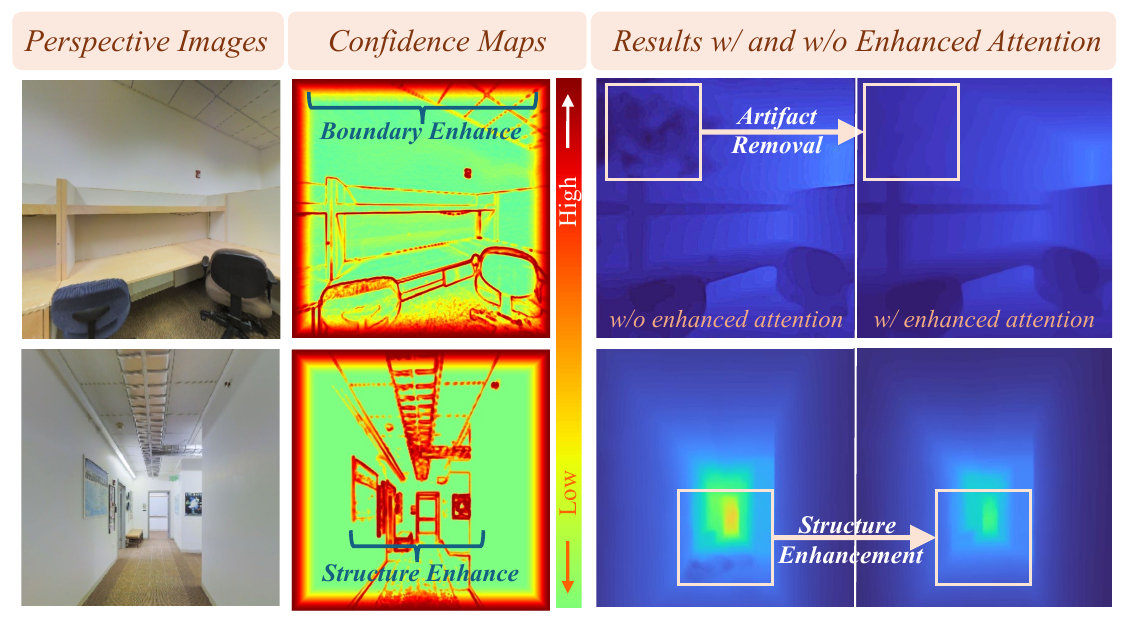}
  \vspace{-6mm}
\caption{
\textbf{Comparison of results before and after applying our \textit{structure-saliency enhanced attention} mechanism}. Guided by our well-designed \textit{structure-aware confidence map}, our VGGT-360 effectively removes artifacts and preserves geometric structures in weakly structured regions, which are easily affected by illumination cues and noise.
}
\vspace{-4mm}
\label{Fig4}
\end{figure}

\subsection{Structure-Saliency Enhanced Attention}
\label{SEA}
Once the perspective view set is obtained, we leverage the powerful 3D-consistent reconstruction capability of VGGT-like foundation models to aggregate multiple perspective views into a globally coherent 3D representation, serving as the foundation for panoramic MDE. However, we observe that these models often deteriorate in weakly structured regions lacking reliable geometric cues, leading to artifacts and hallucinatory depth predictions (Fig.~\ref{Fig4}), despite their strong zero-shot generalization ability. To address this issue, we introduce a \textit{structure-saliency enhanced attention} mechanism that integrates structure-derived priors (\ie, structure-aware confidence maps) into the intra-frame attention layers of VGGT-like models, steering multi-view aggregation toward geometrically stable and reliable regions.

\begin{figure}[t]
  \centering
  \includegraphics[width=1\linewidth]{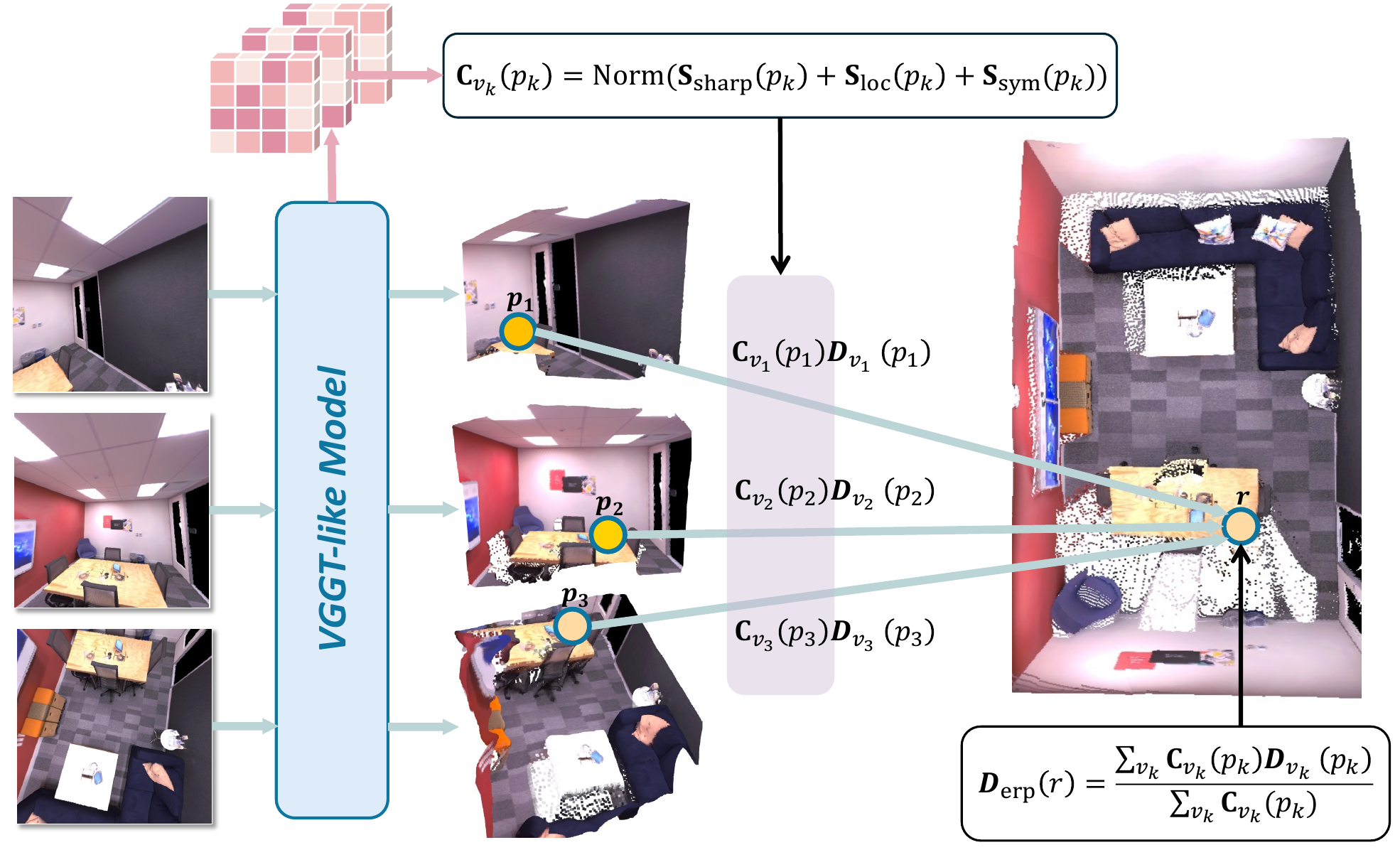}
  \vspace{-6mm}
  \caption{\textbf{Pipeline of our \textit{correlation-weighted 3D model correction} module}. Each overlapping 3D point is assigned a correlation score derived from VGGT’s intra-frame attention. These scores are then used as reliability weights to refine the reconstructed 3D model and produce the final coherent ERP depth.
}
\vspace{-4mm}
\label{Fig6}
\end{figure}

\noindent{\textbf{Structure-Aware Confidence Map.}} 
To inject structural priors into VGGT’s attention without altering its pretrained weights, we define a pixel-level \textit{structure-aware confidence map} $\mathbf{M}_s$ for each view.
As shown in Fig.~\ref{Fig4}, $\mathbf{M}_s$ integrates structural saliency and view boundary-aware weighting to strengthen attention in geometry-reliable regions while preserving feature continuity across overlaps, thereby improving both local depth fidelity and global coherence.

Specifically, we first compute the gradient-based geometric prior $\mathbf{M}_g$ using the Sobel operator 
$\mathbf{Z}(p)$ (see Eq.\ref{eq1}) on the grayscale image to highlight structurally reliable regions: $ \mathbf{M}_g(p){=}\sigma(\mathbf{Z}(p))$.  
To encourage stronger cross-view interaction near overlapping boundaries, we introduce an \textit{edge-band prior} defined as $\mathbf{E}(p)= \mathbf{1}\left[\max(|x|, |y|){\geq}1 - m\right]$,  emphasizing uncertain pixels near the view edges. Here, $(x, y){\in}[-1, 1]^2$ are the normalized image coordinates of pixel $p$, and $m$ controls the edge-band width. The final confidence map is computed as:
\begin{equation}
\mathbf{M}_s(p) = \mathbf{1}_{\mathrm{valid}}(p) \cdot \left[(1 - \mathbf{E}(p)) \cdot \mathbf{M}_g(p) + \mathbf{E}(p)\right].
\end{equation}

\noindent{\textbf{Structure-Saliency Enhanced Frame Attention.}}
\label{sec:confattn}
Given the proposed confidence prior $\mathbf{M}_s$, we enhance intra-frame attention of VGGT-like models to emphasize geometrically reliable keys while suppressing ambiguous ones. 
Let $\mathbf{Q}$, $\mathbf{K}$, and $\mathbf{V}$ denote the query, key, and value features, respectively.  
We incorporate the confidence map $\mathbf{M}_s$ as an additive log-confidence bias to the attention scores:
\begin{equation}
\mathbf{M}_{\mathrm{Attn}} = \operatorname{softmax}\!\left({\mathbf{Q}\mathbf{K}^{\top}}/{\sqrt{d}} + \log(\mathbf{M}_s)\right).
\end{equation}

As shown in Fig.~\ref{Fig4}, this bias explicitly steers attention toward structurally stable regions, while effectively suppressing spurious artifacts in geometrically ambiguous areas, without altering the pretrained backbone weights.

\begin{table*}[h]
\small
\caption{Quantitative comparisons with SOTA methods on Matterport3D~\cite{Matterport3D}, Stanford2D3D~\cite{armeni2017joint}, and Replica360-2K~\cite{straub2019replica, rey2022360monodepth} test sets. \\
}
\centering
\resizebox{0.95\textwidth}{!}{
\begin{tabular}{c|c|c|c|cccc}
\toprule
\textbf{Test Dataset} & \textbf{Method} & \textbf{Backbone} & 
\textbf{Train $\to$ Test} & \textbf{Abs Rel}$\downarrow$ &  $\boldsymbol{\delta_1}\uparrow$ & $\boldsymbol{\delta_2}\uparrow$  & $\boldsymbol{\delta_3}\uparrow$  \\ 
\toprule
\multirow{12}{*}{Matterport3D} & BiFuse++~\cite{wang2022bifuse++} & ResNet34~\cite{he2016deep} & M $\to$ M 
& 0.112   & 0.881  & 0.966 & 0.987  \\
& EGFormer~\cite{yun2023egformer} & Transformer & M $\to$ M 
& 0.147  & 0.816  & 0.939 & 0.974  \\
& HRDFuse~\cite{ai2023hrdfuse} & ResNet34~\cite{he2016deep} & M $\to$ M 
& 0.117   & 0.867  & 0.962 & 0.985  \\
& Elite360D~\cite{ai2024elite360d} & EfficientNet-B5~\cite{tan2019efficientnet} & M $\to$ M 
& 0.105  & 0.899  & 0.971 & \textbf{0.991}  \\
& Depth Anywhere~\cite{wang2024depth} & Depth Anything~\cite{yang2024depth}, UniFuse~\cite{jiang2021unifuse} & M+ $\to$ M 
 & 0.089 & 0.911 & 0.975  & \textbf{0.991} \\ 
 & Depth Anywhere~\cite{wang2024depth} & Depth Anything~\cite{yang2024depth}, BiFuse++~\cite{wang2022bifuse++}  & M+ $\to$ M 
& 0.085  & 0.917  & 0.976 & \textbf{0.991}  \\
\\[-2.3ex] \cline{2-8} \\[-2.3ex]
& 360MD~\cite{rey2022360monodepth}& MiDaS v2~\cite{ranftl2020towards}& Training-free
& 0.264    & 0.612  & 0.854 & 0.941  \\
& RPG360~\cite{jung2025rpg360}& Omnidata v2~\cite{eftekhar2021omnidata}& Training-free
& 0.215   & 0.796  & 0.935 & 0.973  \\
& RPG360~\cite{jung2025rpg360} & Metric3D v2~\cite{hu2024metric3d}& Training-free
& 0.203   & 0.859  & 0.953 & 0.977  \\
&\cellcolor{gray!20}VGGT-360 & \cellcolor{gray!20}VGGT~\cite{wang2025vggt}& \cellcolor{gray!20}Training-free
& \cellcolor{gray!20}0.083   & \cellcolor{gray!20}0.935  & \cellcolor{gray!20}0.978 & \cellcolor{gray!20}0.989  \\
& \cellcolor{gray!20}VGGT-360 & \cellcolor{gray!20}Fastvggt~\cite{shen2025fastvggt}& \cellcolor{gray!20}Training-free
& \cellcolor{gray!20}\textbf{0.078}   & \cellcolor{gray!20}\textbf{0.943}  & \cellcolor{gray!20}\textbf{0.981} & \cellcolor{gray!20}\textbf{0.991}  \\
& \cellcolor{gray!20}VGGT-360 & \cellcolor{gray!20}$\pi^3$~\cite{wang2025pi}& \cellcolor{gray!20}Training-free
& \cellcolor{gray!20}\underline{0.079}  & \cellcolor{gray!20}\underline{0.941}  & \cellcolor{gray!20}\underline{0.979} & \cellcolor{gray!20}0.989  \\
\midrule
\midrule
\multirow{10}{*}{Stanford2D3D} & BiFuse~\cite{wang2020bifuse} & ResNet50~\cite{he2016deep} & M $\to$ S 
& 0.120  & 0.862  & - & - \\
& UniFuse~\cite{jiang2021unifuse} & ResNet34~\cite{tan2019efficientnet} & M $\to$ S 
& 0.094 & 0.913 & - & - \\
& BiFuse++~\cite{wang2022bifuse++} & ResNet34~\cite{he2016deep} & M $\to$ S 
& 0.107  & 0.914  & 0.975 & 0.989 \\
& Depth Anywhere~\cite{wang2024depth} & Depth Anything~\cite{yang2024depth}, UniFuse~\cite{jiang2021unifuse} & M+ $\to$ S 
 & 0.082 & 0.927 & 0.978  & 0.990 \\ 
 & Depth Anywhere~\cite{wang2024depth} & Depth Anything~\cite{yang2024depth}, BiFuse++~\cite{wang2022bifuse++}  & M+ $\to$ S
& 0.083   & 0.930  & 0.978 & 0.990  \\
 & DAC~\cite{guo2025depth} & ResNet101~\cite{he2016deep}  & In+ $\to$ S
&  0.124 & 0.859 & 0.976 & \underline{0.991}  \\
\\[-2.3ex] \cline{2-8} \\[-2.3ex]
& 360MD~\cite{rey2022360monodepth}& MiDaS v2~\cite{ranftl2020towards}& Training-free
& 0.268  & 0.636  & 0.878 & 0.945  \\
& \cellcolor{gray!20}VGGT-360 & \cellcolor{gray!20}VGGT~\cite{wang2025vggt}& \cellcolor{gray!20}Training-free
& \cellcolor{gray!20}\underline{0.068}  & \cellcolor{gray!20}\textbf{0.953}  & \cellcolor{gray!20}\underline{0.983} & \cellcolor{gray!20}\underline{0.991}  \\
& \cellcolor{gray!20}VGGT-360  & \cellcolor{gray!20}Fastvggt~\cite{shen2025fastvggt}& \cellcolor{gray!20}Training-free
&\cellcolor{gray!20}0.070 & \cellcolor{gray!20}\underline{0.952}  &\cellcolor{gray!20}\underline{0.983} & \cellcolor{gray!20}\underline{0.991}  \\
& \cellcolor{gray!20}VGGT-360 & \cellcolor{gray!20}$\pi^3$~\cite{wang2025pi}& \cellcolor{gray!20}Training-free
& \cellcolor{gray!20}\textbf{0.065}  & \cellcolor{gray!20}\underline{0.952}  & \cellcolor{gray!20}\textbf{0.984} & \cellcolor{gray!20}\textbf{0.993}  \\
\midrule
\midrule
\multirow{12}{*}{Replica360-2K} & BiFuse~\cite{wang2020bifuse} & ResNet50~\cite{he2016deep} & M $\to$ R 
& 0.318   & 0.591  & 0.840 & 0.927\\
& HoHoNet~\cite{sun2021hohonet} & ResNet50~\cite{he2016deep} & M $\to$ R 
& 0.259   & 0.672  & 0.888 & 0.942\\ 
& UniFuse~\cite{jiang2021unifuse} & ResNet34~\cite{he2016deep} & M $\to$ R
& 0.233 & 0.728  & 0.905 & 0.954\\
& Depth Anywhere~\cite{wang2024depth} & Depth Anything~\cite{yang2024depth}, UniFuse~\cite{jiang2021unifuse} & M+ $\to$ R 
 & 0.219  & 0.763 & 0.905  & 0.951 \\ 
 & Depth Anywhere~\cite{wang2024depth} & Depth Anything~\cite{yang2024depth}, BiFuse++~\cite{wang2022bifuse++}  & M+ $\to$ R
& 0.223   & 0.758  & 0.903 & 0.955  \\
 & DAC~\cite{guo2025depth} & ResNet101~\cite{he2016deep}  & In+ $\to$ R
& 0.142  & 0.803 & 0.960 & 0.994  \\
\\[-2.3ex] \cline{2-8} \\[-2.3ex]
& 360MD~\cite{rey2022360monodepth}& MiDaS v2~\cite{ranftl2020towards}& Training-free
& 0.167  & 0.769  & 0.948 & 0.983  \\
& HDE360~\cite{peng2023high}& UniFuse~\cite{jiang2021unifuse}& Training-free
& 0.133  & 0.870  & 0.961 & 0.978  \\
& HDE360~\cite{peng2023high}& HoHoNet~\cite{sun2021hohonet}& Training-free
& 0.107  & 0.910  & 0.961 & 0.982  \\
& \cellcolor{gray!20}VGGT-360  & \cellcolor{gray!20}VGGT~\cite{wang2025vggt}& \cellcolor{gray!20}Training-free
& \cellcolor{gray!20}0.075 & \cellcolor{gray!20}0.934 & \cellcolor{gray!20}0.985 & \cellcolor{gray!20}0.993 \\
& \cellcolor{gray!20}VGGT-360  & \cellcolor{gray!20}Fastvggt~\cite{shen2025fastvggt}& \cellcolor{gray!20}Training-free
& \cellcolor{gray!20}\textbf{0.069} & \cellcolor{gray!20}\textbf{0.950} & \cellcolor{gray!20}\textbf{0.989} & \cellcolor{gray!20}\textbf{0.996}  \\
& \cellcolor{gray!20}VGGT-360  & \cellcolor{gray!20}$\pi^3$~\cite{wang2025pi}& \cellcolor{gray!20}Training-free
& \cellcolor{gray!20}\underline{0.070} & \cellcolor{gray!20}\underline{0.941} & \cellcolor{gray!20}\textbf{0.989} & \cellcolor{gray!20}\textbf{0.996}  \\
\bottomrule
\end{tabular}}
\vspace{2mm}

\small
\textbf{M}: Matterport3D~\cite{Matterport3D}  \hspace{1.5em}
\textbf{M+}: Matterport3D~\cite{Matterport3D} + pseudo-labeled Structured3D~\cite{zheng2020structured3d}
\hspace{1.5em}
\textbf{In+}: Joint indoor datasets~\cite{ramakrishnan2021habitat, zamir2018taskonomy, roberts2021hypersim}
\label{tab1} \end{table*}

\subsection{Correlation-Weighted 3D Model Correction}
\label{CMC}
Building on the reconstructed 3D representation in Sec.~\ref{SEA}, we introduce a \textit{correlation-weighted 3D model correction} module to further enforce geometric consistency across overlapping regions, as shown in Fig.~\ref{Fig6}. Our core idea is to assign each overlapping 3D point a correlation score, derived from intra-frame attention, that reflects its connectivity with neighboring points. Points with higher correlations are regarded as more reliable and thus receive greater weights for the final ERP depth construction.

Specifically, consider an ERP pixel $r$ that is observed by a set of $N_{K}$ perspective views, denoted as $\mathcal{V}^k_{\mathrm{per}}{=}\{v_1, \dots, v_{N_{K}}\}$. 
For each view $v_k$, we project the ERP pixel $r$ onto its corresponding 3D point $p_k$ to obtain the depth observation $\mathcal{D}_{v_k}(p_k)$. Subsequently, a correlation weight $\mathbf{C}_{v_k}(p_k)$ for the point $p_k$ is derived from that view's attention map $\widetilde{\mathbf{M}}_{\mathrm{Attn}}$ of the final intra-frame attention layer. Finally, the final ERP depth $\mathcal{D}_{\mathrm{erp}}$ is computed via a pixel-wise weighted aggregation of these multi-view depths:
\begin{equation}
\mathcal{D}_{\mathrm{erp}}(r) =
\frac{\sum_{v_k} \mathbf{C}_{v_k}(p_k) \mathcal{D}_{v_k}(p_k)}
{\sum_{v_k} \mathbf{C}_{v_k}(p_k)}.
\end{equation}

\begin{figure}[t]
  \centering
  \includegraphics[width=0.95\linewidth]{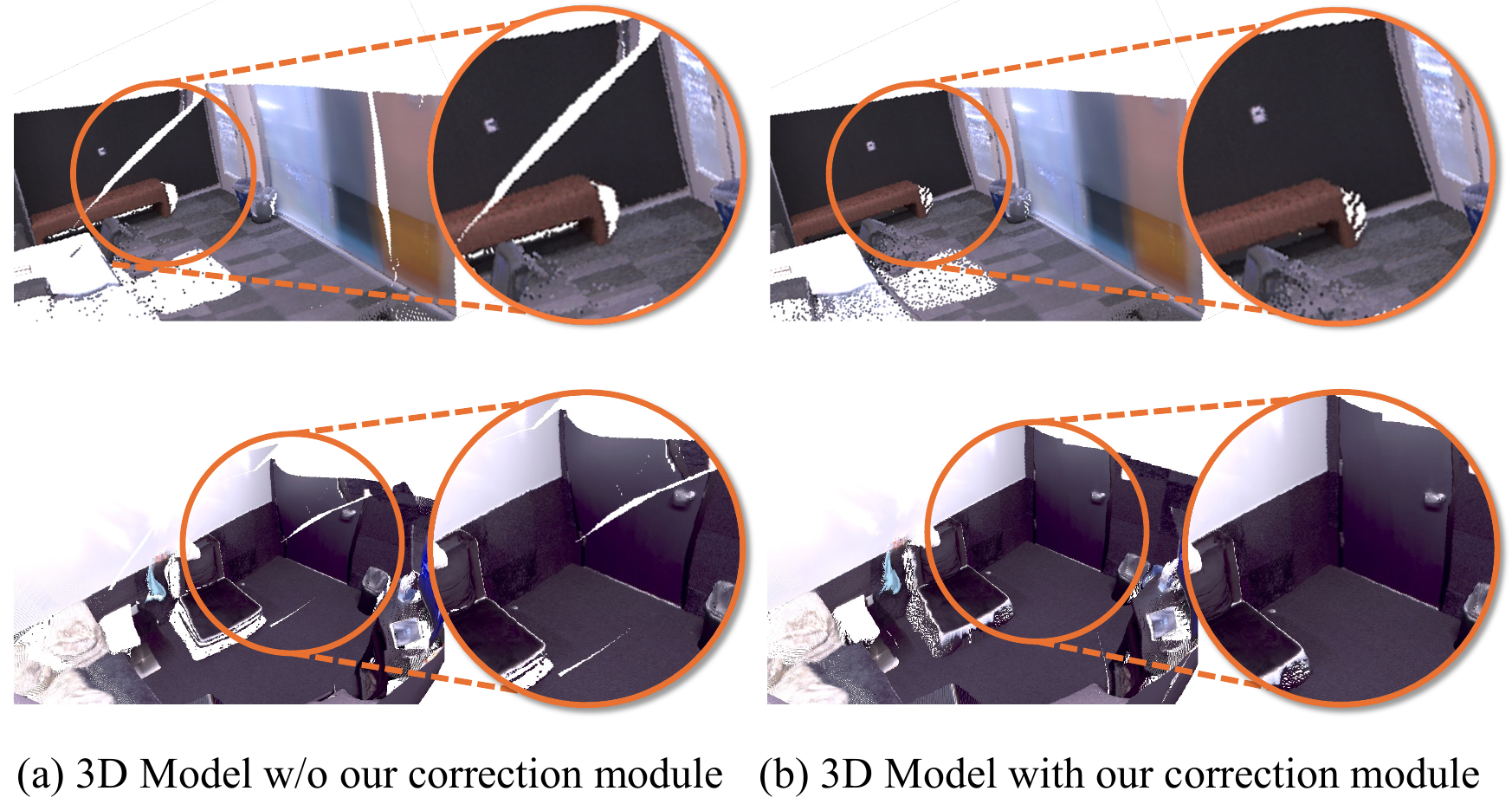}
  \caption{
  \textbf{Comparison of reconstructed 3D models without (a) and with (b) our \textit{correlation-weighted 3D model correction}}. Our correction module significantly enhances surface continuity and removes artifacts in overlapping regions.
}
\vspace{-4mm}
\label{Fig5}
\end{figure}

To derive the correlation weight $\mathbf{C}_{v_k}(p_k)$, we compute three complementary measures: \textit{sharpness}, \textit{locality}, and \textit{symmetry}, from the attention map $\widetilde{\mathbf{M}}_{\mathrm{Attn}}$. 

\begin{itemize}
\item \textbf{Sharpness.} $\mathbf{S}_{\mathrm{sharp}}(p_k)$ quantifies the concentration of attention around the point $p_k$.
A sharply peaked distribution indicates that the model focuses confidently on a few stable correspondences, implying stronger geometric reliability.
We compute it using the normalized Shannon entropy, where lower entropy corresponds to sharper attention and thus higher $\mathbf{S}_{\mathrm{sharp}}(p_k)$:
\begin{equation}
\small
\begin{gathered}
\mathbf{S}_{\mathrm{sharp}}(p_k)
= 1 - \frac{\operatorname{H}(p_k)}{\log |\Omega(v_k)|},\\
\operatorname{H}(p_k)
= - \sum_{p \in \Omega(v_k)}
\widetilde{\mathbf{M}}_{\mathrm{Attn}}(p_k, p)
\log \widetilde{\mathbf{M}}_{\mathrm{Attn}}(p_k, p),
\end{gathered}
\end{equation}
where $\Omega(v_k)$ denotes the set of points within view $v_k$.
\item \textbf{Locality.} $\mathbf{S}_{\mathrm{loc}}(p_k)$ measures the spatial compactness of attention around the point $p_k$, as stable geometric regions typically attend locally, while long-range attention suggests unreliable local features being compensated by distant cues.
We compute $\mathbf{S}_{\mathrm{loc}}(p_k)$ by weighting spatial distances with a Gaussian kernel $G(\cdot)$:
\begin{equation}
\small
\mathbf{S}_{\mathrm{loc}}(p_k)
= \sum_{p \in \Omega(v_k)}
\widetilde{\mathbf{M}}_{\mathrm{Attn}}(p_k, p)\,
G\!\big(\|\mathbf{x}_p - \mathbf{x}_{p_k}\|\big),
\end{equation}
where $\mathbf{x}_{p_k}$ and $\mathbf{x}_p$ denote 2D coordinates of $p_k$ and $p$.
\item \textbf{Symmetry.} $\mathbf{S}_{\mathrm{sym}}(p_k)$ quantifies the mutual consistency of attention around the point $p_k$, as reliable geometric correspondences are typically bidirectional (\ie, if $p_k$ attends strongly to another point $p$, $p$ should also attend back to $p_k$).
We compute it using the Bhattacharyya coefficient:
\begin{equation}
\small
\mathbf{S}_{\mathrm{sym}}(p_k)
= \sum_{u \in \Omega(v_k)}
\sqrt{\,
\widetilde{\mathbf{M}}_{\mathrm{Attn}}(p_k, p)\,
\widetilde{\mathbf{M}}'_{\mathrm{Attn}}(p_k, p)
\,},
\end{equation}
where $\widetilde{\mathbf{M}}'_{\mathrm{Attn}}$ is the normalized transpose of $\widetilde{\mathbf{M}}_{\mathrm{Attn}}$.
\end{itemize}
The three correlation metrics are additively aggregated and then normalized to derive the correlation weight $\mathbf{C}_{v_k}(p_k)$:
\begin{equation}
\small
\mathbf{C}_{v_k}(p_k) = \operatorname{Norm}\big( \mathbf{S}_{\mathrm{sharp}}(p_k) + \mathbf{S}_{\mathrm{loc}}(p_k) + \mathbf{S}_{\mathrm{sym}}(p_k) \big),
\end{equation}
where all metrics are pre-normalized to ensure balanced contributions. As shown in Fig.~\ref{Fig5}, the refined 3D model with our correction model demonstrates significantly improved geometric coherence and higher accuracy.

\section{Experiment}

\begin{figure*}[t]
  \centering
  \includegraphics[width=1\linewidth]{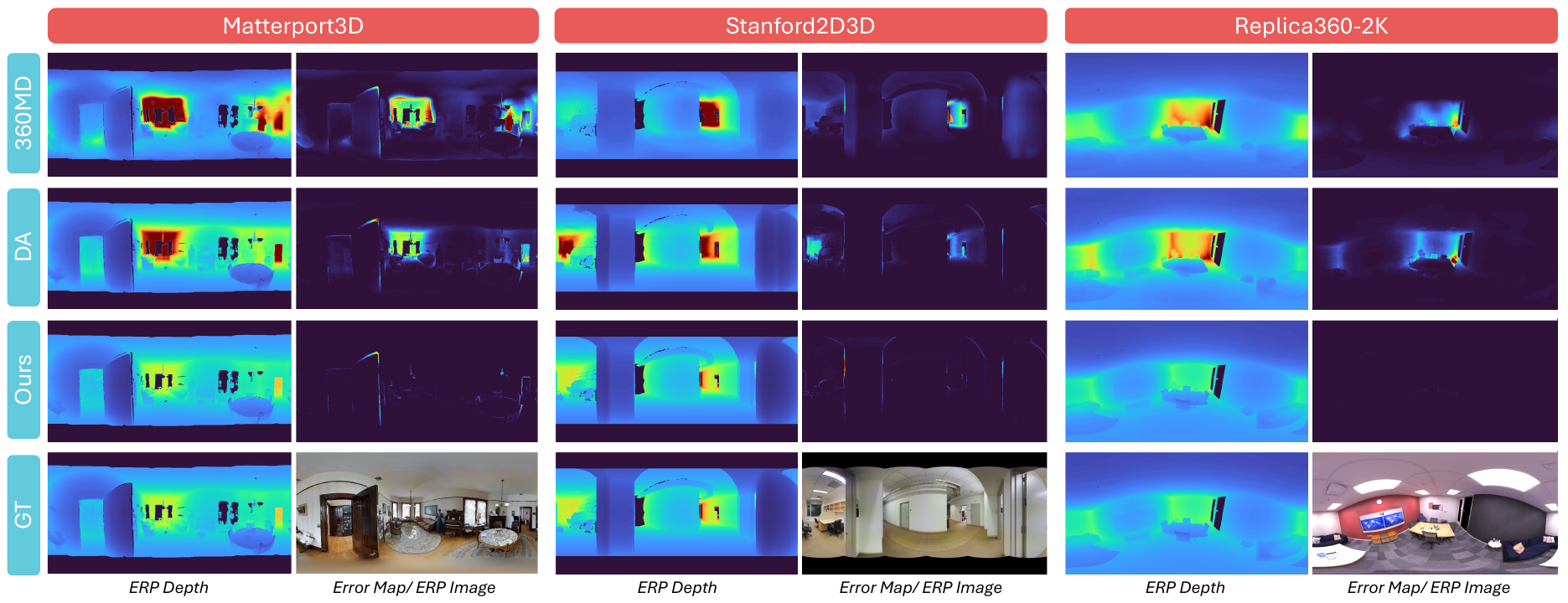}
  \vspace{-7mm}
  \caption{Qualitative comparisons with the state-of-the-art supervised method Depth-Anywhere (DA)~\cite{wang2024depth} and the training-free method 360MD~\cite{rey2022360monodepth} across three indoor datasets: Matterport3D~\cite{Matterport3D}, Stanford2D3D~\cite{armeni2017joint}, and Replica360-2K~\cite{straub2019replica, rey2022360monodepth}.
}
  \vspace{-2mm}
\label{Indoor_results}
\end{figure*}

\noindent\textbf{Datasets.}
For benchmarking, we evaluate on three standard indoor datasets: \textbf{Matterport3D}~\cite{Matterport3D}, \textbf{Stanford2D3D}~\cite{armeni2017joint}, and \textbf{Replica360-2K}~\cite{straub2019replica, rey2022360monodepth}, which provide groundtruth depth and cover diverse resolutions and indoor layouts.
To assess generalization under outdoor scenarios, following 360MD~\cite{rey2022360monodepth}, we further provide qualitative results from \textbf{OmniPhotos}~\cite{bertel2020omniphotos}, where groundtruth is unavailable.

\noindent\textbf{Implementation.}
During the adaptive projection phase, we use $N_{\mathcal{B}}{=}8$ base views and augment the top-$K{=}2$ most uncertain ones.
For enhanced attention, we set $m{=}0.05$ to modulate the sharpness and border width of the structure-aware confidence map. We evaluate our framework on three VGGT-like baselines: VGGT~\cite{wang2025vggt}, $\pi^3$~\cite{wang2025pi}, and Fastvggt~\cite{shen2025fastvggt}.
All experiments are conducted on NVIDIA RTX 4090 and TITAN RTX GPUs.

\noindent\textbf{Baselines.} We compare our VGGT-360 with \textbf{three} state-of-the-art (SOTA) \textit{training-free} panoramic depth estimation methods: 360MD~\cite{rey2022360monodepth}, HDE360~\cite{peng2023high}, and RPG360~\cite{jung2025rpg360}, as well as \textbf{nine} SOTA \textit{fully-trained} methods: BiFuse~\cite{wang2020bifuse}, UniFuse~\cite{jiang2021unifuse}, HoHoNet~\cite{sun2021hohonet}, BiFuse++~\cite{wang2022bifuse++}, EGFormer~\cite{yun2023egformer}, HRDFuse~\cite{ai2023hrdfuse}, Elite360D~\cite{ai2024elite360d}, Depth Anywhere~\cite{wang2024depth}, and DAC~\cite{guo2025depth}.
All fully-trained methods are trained on the Matterport3D training set (\textbf{M}), except Depth Anywhere~\cite{wang2024depth}, which additionally leverages Structured3D~\cite{zheng2020structured3d} pseudo-labels (\textbf{M+}), and DAC~\cite{guo2025depth}, which is trained jointly on three indoor datasets~\cite{ramakrishnan2021habitat, zamir2018taskonomy, roberts2021hypersim} (\textbf{In+}).

\subsection{Comparison with SOTA methods}
Table~\ref{tab1} reports quantitative comparisons with SOTA methods across three benchmarks. On \textbf{Matterport3D}~\cite{Matterport3D}, although all supervised baselines are trained on its official set, VGGT-360 still achieves the best overall performance, outperforming both training-free and fully-trained methods. Notably, it even surpasses Depth Anywhere~\cite{wang2024depth}, which is trained on large-scale labeled data and leverages powerful pretrained large depth models. This result highlights the effectiveness of our geometry-aware VGGT-360 framework, which achieves consistent and structure-preserving depth without any task-specific adaptation.

We further assess the zero-shot generalization of all methods on \textbf{Stanford2D3D}~\cite{armeni2017joint}, a challenging benchmark with domain shifts in appearance and layout. While supervised models often rely on dataset-specific priors and texture cues, our VGGT-360 achieves high accuracy by enforcing view-consistent geometric reasoning. This highlights the inherent generalizability of our geometry-grounded design across diverse and unseen environments.

We also conduct zero-shot evaluation on the more challenging \textbf{Replica360-2K}~\cite{straub2019replica, rey2022360monodepth}, which contains high-resolution indoor scenes. Our VGGT-360 still achieves the best overall performance, demonstrating strong scalability to high-resolution inputs. This robustness is attributed to our adaptive projection and multi-view geometric reasoning, which together ensure global consistency and structural fidelity across varying scene complexities and resolutions.

\begin{figure}[t]
  \centering

  \includegraphics[width=1\linewidth]{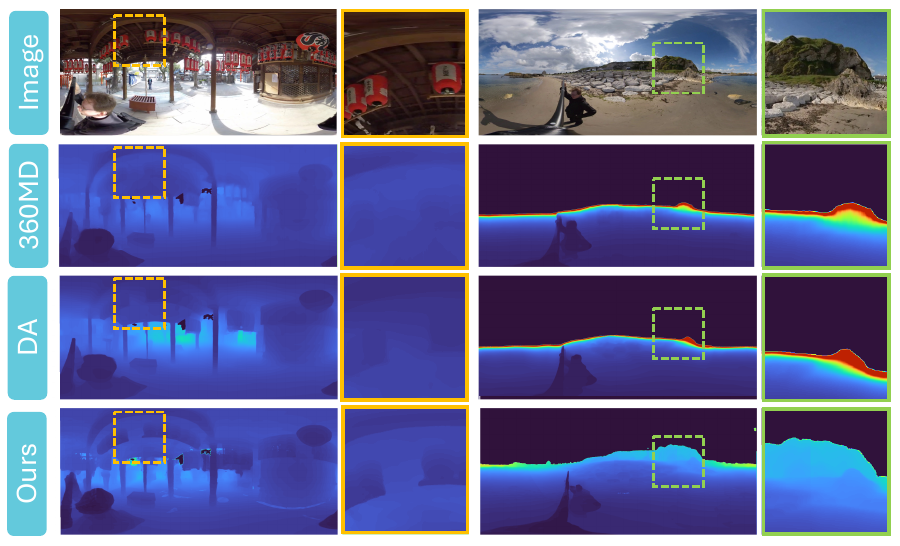}
    \vspace{-5mm}
  \caption{Qualitative comparison with SOTA methods: supervised Depth-Anywhere (DA)~\cite{wang2024depth} and training-free 360MD~\cite{rey2022360monodepth}, on outdoor panoramas from OmniPhotos~\cite{bertel2020omniphotos}.
   \vspace{-4mm}
}
\label{outdoor_results}
\end{figure}

\subsection{Visualization on Indoor and Outdoor Scenes}
Figure~\ref{Indoor_results} presents qualitative comparisons on \textbf{three indoor datasets}: Matterport3D~\cite{Matterport3D}, Stanford2D3D~\cite{armeni2017joint}, and Replica360-2K~\cite{straub2019replica, rey2022360monodepth}. We compare our method against a fully-supervised baseline (Depth Anywhere~\cite{wang2024depth}) and a training-free baseline (360MD~\cite{rey2022360monodepth}).
Across all scenes, VGGT-360 produces more accurate and globally coherent depth maps, effectively preserving object boundaries and suppressing depth inconsistencies. The advantages of our method are especially evident under zero-shot and high-resolution settings, where baseline approaches often suffer from over-smoothing and inaccurate geometric predictions. In contrast, our globally consistent pipeline captures scene structure faithfully without training. The accompanying error maps further confirm the superiority of our predictions, showing fewer high-error regions, especially in geometrically challenging areas such as distant surfaces.
\begin{figure}[t]
  \centering
\includegraphics[width=0.9\linewidth]{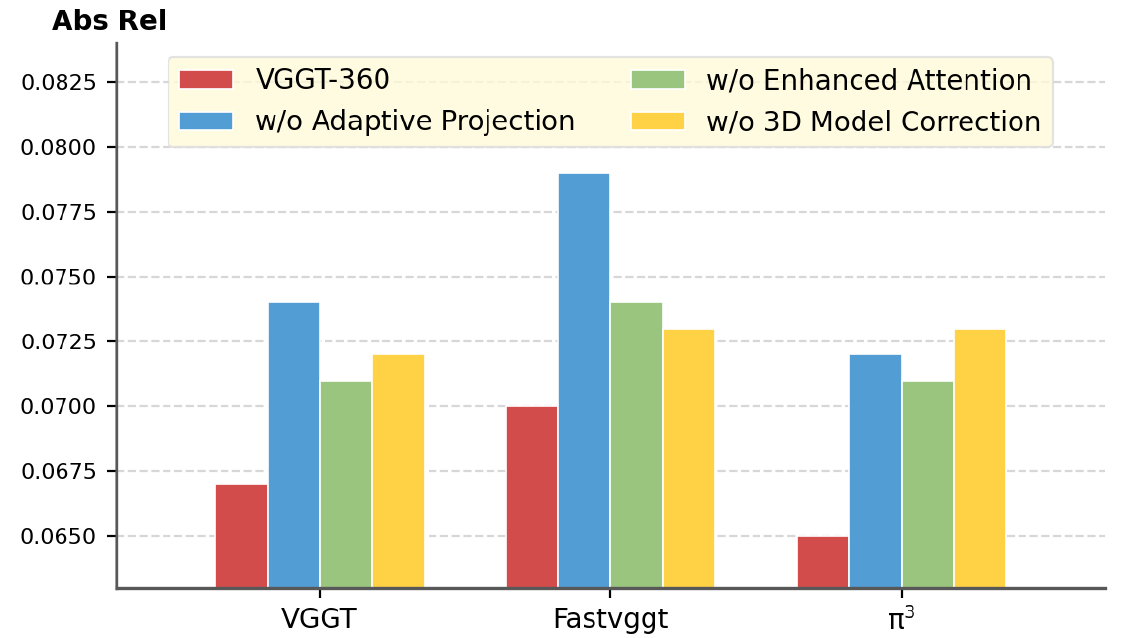}
\vspace{-2mm}
  \caption{Ablation studies on Stanford2D3D~\cite{armeni2017joint}: effectiveness of the three proposed modules across various VGGT-like baselines.
}
\vspace{-4mm}
\label{ab_fig1}
\end{figure}

Figure~\ref{outdoor_results} presents qualitative comparisons on \textbf{outdoor panoramas} from OmniPhotos~\cite{bertel2020omniphotos}, which features diverse real-world scenes with complex geometry, rich textures, and wide depth ranges.
Since no ground-truth depth is available, following 360MD~\cite{rey2022360monodepth}, we compare methods based on visual plausibility. Our VGGT-360 yields sharper and more structure-preserving predictions, retaining fine architectural details and geometric consistency. 
These results highlight the strong zero-shot capability of VGGT-360 in unconstrained outdoor environments, demonstrating its robustness and practicality across diverse panoramic domains.

\subsection{Ablation Study}

\noindent\textbf{Generalization and Effectiveness of Each Module.} As shown in Fig.~\ref{ab_fig1}, all three proposed modules consistently improve accuracy across diverse VGGT-like backbones, validating the strong generality of our design. Each module operates without retraining, exhibits high plug-and-play compatibility across architectures, and collectively delivers robust gains in panoramic depth estimation.

\begin{figure}[t]
  \centering
\includegraphics[width=1\linewidth]
{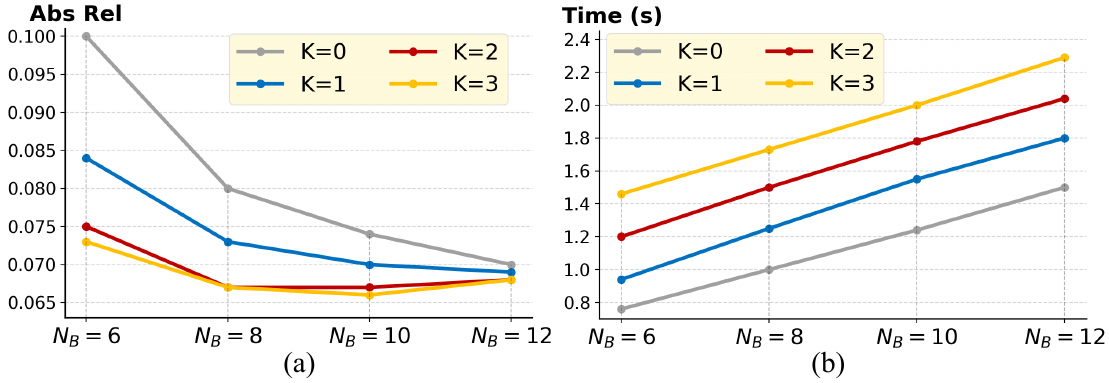}
\vspace{-7mm}
  \caption{Ablation studies on Stanford2D3D~\cite{armeni2017joint}: impact of projection parameters (base views ($N_\mathcal{B}{=}6$-$12$) and top-$K$ neighbor augmentation ($K{=}0$-$3$)) on performance (a) and runtime (b).
}
\vspace{-4mm}
\label{ab_fig}
\end{figure}

\noindent\textbf{Impact of Different Projection Strategies.}
Building on the above analysis, we conduct ablation on the projection module using Stanford2D3D~\cite{armeni2017joint}, varying the number of base and adaptively selected neighbor views. As shown in Fig.~\ref{ab_fig}, uniform projection with $N_\mathcal{B}{=}6$ views offers limited performance, while simply increasing the view count brings marginal gains at a high computational cost. In contrast, our adaptive strategy (\ie, $N_\mathcal{B}{=}8$ base views with top-$K{=}2$ neighbor augmentation) achieves a better accuracy–efficiency trade-off, showing that dynamically focusing on uncertain regions outperforms fixed sampling.

\begin{table}[t]
\vspace{-2mm}
\caption{Ablation of VGGT~\cite{wang2025vggt} baseline on Stanford2D3D~\cite{armeni2017joint}.}
\vspace{-1mm}
\centering
\small
\resizebox{0.95\columnwidth}{!}{
\begin{tabular}{l c c c}
\toprule
\textbf{Method} & \textbf{Abs Rel}$\downarrow$ & \textbf{RMSE}$\downarrow$ & \textbf{Time} \\
\midrule
\multicolumn{4}{l}{\textit{Effect of Structure-Saliency Enhanced Attention}} \\
\midrule
Baseline & 0.080 & 0.354 & \textbf{1.41}s \\
Baseline + $\mathbf{M}_{g}$ & 0.075 & 0.343 & 1.43s \\
Baseline + $\mathbf{M}_{g}$ + $\mathbf{E}$ & 0.073 & 0.346 & 1.44s \\
Baseline + $\mathbf{M}_{g}$+ $\mathbf{E}$ + $\mathbf{1}_{\mathrm{valid}}$ & \textbf{0.072} & \textbf{0.340} & 1.45s\\
\midrule
\multicolumn{4}{l}{\textit{Effect of Correlation-Weighted 3D Model Correction}} \\
\midrule
Baseline & 0.080 & 0.354 & \textbf{1.41}s \\
Baseline + $\mathbf{S}_\mathrm{sharp}$ & 0.074 &0.328  & 1.43s \\
Baseline + $\mathbf{S}_{\mathrm{loc}}$ & 0.073 & 0.327 & 1.44s \\
Baseline + $\mathbf{S}_\mathrm{Sym}$ & 0.074 & 0.328 & 1.44s \\
Baseline + $\mathbf{S}_\mathrm{sharp} + \mathbf{S}_\mathrm{loc} + \mathbf{S}_\mathrm{Sym}$ & \textbf{0.071} & \textbf{0.325} & 1.46s  \\
\bottomrule
\end{tabular}}
\label{tab_ab}
\end{table}

\noindent\textbf{Effect of Structure-Saliency Enhanced Attention.} As shown in Table~\ref{tab_ab}, we analyze the effect of each component in the proposed Structure-Saliency Enhanced Attention. Incorporating the gradient-based saliency prior $\mathbf{M}_{g}$ improves accuracy by emphasizing structural edges. Adding the edge-band prior $\mathbf{E}$ further enhances boundary regions, while the validity mask $\mathbf{1}_{\mathrm{valid}}$ yields the best results, showing the complementary roles of saliency, edge weighting, and view validity in refining attention.

\noindent\textbf{Effect of Correlation-Weighted 3D Model Correction.}
We further ablate each component of the correlation-weighted 3D model correction module, as shown in Table~\ref{tab_ab}. Starting from the baseline, adding any single correlation cue (\ie, sharpness $\mathbf{S}_\mathrm{sharp}$, locality $\mathbf{S}_\mathrm{loc}$, or symmetry $\mathbf{S}_\mathrm{sym}$) improves accuracy. Combining all three yields the best performance, showing that their synergy promotes more reliable 3D structures and higher-quality depth.

\section{Conclusion}
In this paper, we presented VGGT-360, a training-free framework for panoramic depth estimation that reformulated the task as panoramic reprojection from a globally consistent 3D model. Leveraging VGGT’s intrinsic 3D reasoning, our method overcame the limitations of prior view-independent methods without any labels or training.
We introduced three key modules: 1) uncertainty-guided adaptive projection for geometry-aware view sampling, 2) structure-saliency enhanced attention for robust 3D reconstruction, and 3) correlation-weighted 3D model correction for reliable ERP depth estimation.
Experiments showed VGGT-360 surpassed prior training-free and supervised methods, demonstrating robust zero-shot performance.

\section{Acknowledgments}
This research was supported by the Ministry of Education, Singapore, under its MOE Academic Research Fund Tier 2 (MOE-T2EP20124-0013).
{
    \small
    \bibliographystyle{ieeenat_fullname}
    \bibliography{main}
}


\end{document}